\documentclass[conference]{IEEEtran}
\IEEEoverridecommandlockouts

\usepackage{cite}
\usepackage{amsmath,amssymb,amsfonts}
\usepackage{algorithmic}
\usepackage{graphicx}
\usepackage{textcomp}
\usepackage{xcolor}
\usepackage{booktabs}
\usepackage{siunitx}
\usepackage{multirow}
\usepackage{url}
\usepackage{hyperref}
\usepackage{enumitem}
\hypersetup{hidelinks}
\usepackage[caption=false]{subfig}
\def\BibTeX{{\rm B\kern-.05em{\sc i\kern-.025em b}\kern-.08em
    T\kern-.1667em\lower.7ex\hbox{E}\kern-.125emX}}

\begin{document}

\title{Enhancing Diffusion Policy with Classifier-Free Guidance for Temporal Robotic Tasks\\}

\author{
\IEEEauthorblockN{Yuang Lu, Song Wang\IEEEauthorrefmark{1}, Xiao Han,
Xuri Zhang, Yucong Wu, Zhicheng He}
\IEEEauthorblockA{Leju (Suzhou) Robot Technology Co., Ltd\\
Jiangsu, China \\
\{luyuang, wangsong, hanxiao, zhangxuri, wyc, hzc\}@lejurobot.com}

}

\maketitle

\begin{abstract}
Temporal sequential tasks challenge humanoid robots, as existing Diffusion Policy (DP) and Action Chunking with Transformers (ACT) methods often lack temporal context, resulting in local optima traps and excessive repetitive actions. To address these issues, this paper introduces a Classifier-Free
Guidance-Based Diffusion Policy (CFG-DP), a novel framework to enhance DP by integrating Classifier-Free Guidance (CFG) with conditional and unconditional models. Specifically, CFG leverages timestep inputs to track task progression and ensure precise cycle termination. It dynamically adjusts action predictions based on task phase, using a guidance factor tuned to balance temporal coherence and action accuracy. Real-world experiments on a humanoid robot demonstrate high success rates and minimal repetitive actions. Furthermore, we assessed the model's ability to terminate actions and examined how different components and parameter adjustments affect its performance. This framework significantly enhances deterministic control and execution reliability for sequential robotic tasks.
\end{abstract}

\begin{IEEEkeywords}
humanoid robot, diffusion policy, classifier-free guidance, robot control, imitation learning
\end{IEEEkeywords}

\section{Introduction}
Diffusion-based policies, such as Diffusion Policy (DP)~\cite{b1} and Action Chunking with Transformers (ACT)~\cite{b2}, have shown remarkable success in robotic manipulation by leveraging denoising diffusion probabilistic models (DDPMs) to generate action sequences from sensory inputs. These methods excel in tasks requiring multimodal action distributions, such as pushing or pouring, achieving significant performance improvements over traditional approaches~\cite{b1}. Researchers are actively exploring ways to endow humanoid robots with generalizable intelligence for diverse tasks, and enabling robots to perform diverse tasks has emerged as a critical focus in artificial intelligence, robotics, and human-computer interaction~\cite{b6}.

Temporal sequential tasks with repetitive cycles and precise termination requirements pose significant challenges for existing control frameworks like DP and ACT. These methods map sensory inputs---typically single-frame camera data and motor states---to action sequences but struggle to accurately judge when to terminate a task, especially in dynamic environments. Their reliance on short-term, single-frame inputs fails to capture the broader temporal context necessary for distinguishing between continuing a cycle and concluding a task, often leading to repetitive actions and suboptimal performance~\cite{b3}. Moreover, DP and ACT employ diffusion models with implicit action distributions, lacking explicit probability outputs, which introduces uncertainty at critical decision points, particularly in tasks with periodic states where cycle completion is ambiguous~\cite{b4,b7}. In practical applications, such as robotic imitation of human screw-tightening motions, robots may persist in redundant rotations instead of retracting the arm due to insufficient cues about task progress. 

To address these limitations, we propose a novel framework that enhances termination judgment by integrating Classifier-Free Guidance (CFG)~\cite{b22,b23}, as applied in goal-conditioned imitation learning~\cite{b5}, training conditional and unconditional models with a dynamic guidance factor \(\lambda\). Additionally, we model timestep counts as an input, enabling explicit tracking of task progression to ensure precise cycle completion.

Our model’s design draws inspiration from human intuition in repetitive tasks, where individuals naturally monitor cycle progress to determine when to conclude actions, such as stopping a screwing motion after a set number of rotations. We translate this insight into a technical framework by incorporating timestep counts as a temporal cue, mirroring human-like awareness of task progression to prevent repetitive loops. Building on the BEhavior generation with ScOre-based Diffusion Policies (BESO) framework~\cite{b5}, which uses Classifier-Free Guidance (CFG) for goal-conditioned tasks, our approach simplifies the model for cyclic manipulation by relying on lightweight step-based guidance, achieving robust termination with lower computational cost than BESO’s multi-task focus. This human-inspired design directly addresses the repetitive action problem, paving the way for our proposed solution~\cite{b5}.

The motivation for this work stems from the need to achieve deterministic control and reliable execution in temporal sequential tasks, such as iterative robotic manipulation processes with distinct cycle endpoints~\cite{b3}. Deterministic control is essential for robots to predictably navigate task phases, ensuring precise transitions to termination actions without indefinite repetitions, a challenge unmet by DP and ACT due to their lack of temporal awareness~\cite{b1,b2}. Equally critical is improving task execution reliability, as uncertain action selection in cyclic tasks can lead to failures, undermining dependability in high-stakes applications~\cite{b6,b4}. By leveraging CFG and explicit temporal modeling, our approach aims to enhance the robustness and applicability of robotic systems, enabling confident and consistent performance in complex, time-dependent tasks.

\section{Related Work}
The development of robust control policies for sequential robotic tasks has been a focal point in robotic manipulation research, with recent advancements leveraging generative models like diffusion policies to address complex, multimodal behaviors~\cite{b1,b2}. However, challenges in modeling temporal progression and accurately judging termination conditions remain critical barriers to achieving deterministic and reliable task execution. Below, we review key works related to incorporating temporal inputs into diffusion-based policies and improving termination condition judgment in robotic tasks, highlighting their contributions and limitations in the context of our proposed framework.

\subsection{Temporal Modeling in Diffusion Policies}
DP and ACT have demonstrated significant success in visuomotor policy learning by mapping sensory inputs to action sequences. However, their reliance on instantaneous inputs limits their ability to capture temporal context, often leading to suboptimal decisions in tasks requiring cycle completion awareness~\cite{b3}. To address this, recent studies have explored explicit temporal modeling. Høeg et al.~\cite{b9} proposed the Temporally Entangled Diffusion (TEDi) Policy, which incorporates temporal step inputs to model long-range dependencies in robotic control tasks. By conditioning the diffusion process on a temporal step variable, TEDi enhances action sequence coherence, reducing myopic planning. However, TEDi focuses on navigation tasks and does not explicitly address termination conditions, limiting its applicability to manipulation tasks with distinct cycle endpoints.

Similarly, Hu et al.~\cite{b10} introduced temporal condition guidance in the Instructed Diffuser framework for offline reinforcement learning, where a step-based temporal input guides the diffusion model to align actions with task progress. While effective for long-horizon tasks, this approach assumes predefined reward signals, which are often unavailable in imitation learning settings like ours. Ge et al.~\cite{b11} explored temporal step inputs in the Spatiotemporal Predictive Pre-training (STP) framework, using paired video frames to predict motion dynamics. STP’s focus on pre-training visual representations improves spatial-temporal feature learning but does not directly address policy-level termination decisions. Other recent works have also explored temporal modeling in diffusion policies for robotic tasks. For instance, RDT-1B~\cite{b16} presents a diffusion foundation model specifically designed for bimanual manipulation, incorporating temporal dependencies to handle complex action sequences. PlayFusion~\cite{b17} leverages diffusion models for skill acquisition from language-annotated play data, emphasizing temporal coherence in skill learning. Additionally, Diffusion-VLA~\cite{b18} combines diffusion and autoregressive modeling to scale robot foundation models, enhancing their ability to manage long-term temporal dependencies. These works highlight the potential of temporal inputs but fall short of providing mechanisms for explicit task progression modeling or termination judgment, key requirements for our sequential task framework.

\subsection{Termination Condition Judgment in Robotic Tasks}
Accurately judging termination conditions is critical for sequential robotic tasks, where premature or delayed termination can lead to task failures or repetitive actions~\cite{b3}. Traditional DP and ACT frameworks struggle with this due to their implicit action distributions, which lack explicit probability outputs for decision-making~\cite{b4}. Recent efforts have sought to address this challenge. Ze et al.~\cite{b12} proposed the 3D Diffusion Policy, which enhances termination judgment by incorporating 3D point cloud inputs and a receding horizon control mechanism. While effective for multi-task manipulation, its focus on spatial robustness does not explicitly tackle temporal termination cues, limiting its ability to handle cyclic tasks with periodic states.

Another approach is the use of Signal Temporal Logic (STL) to enforce rule-compliant behaviors, as explored by Meng and Fan~\cite{b13}. Their Diverse Controllable Diffusion Policy integrates STL with diffusion models to ensure tasks adhere to temporal constraints, such as completing a cycle within a specified timeframe. This method generates diverse, rule-aware trajectories but relies on predefined logical specifications, which may be impractical for uncurated datasets typical in imitation learning. Mishra et al.~\cite{b14} introduced Generative Skill Chaining, using diffusion models to plan long-horizon skills with termination conditions modeled as keypose transitions. While effective for skill sequencing, this approach assumes discrete skill boundaries, which may not generalize to continuous, cyclic tasks like those in our study. 

CFG, as proposed by Reuss et al.~\cite{b5}, offers a promising avenue for improving termination judgment. By training conditional and unconditional models, CFG enhances goal-directed behavior in goal-conditioned imitation learning, allowing policies to prioritize termination actions when appropriate. However, its application to step-based temporal tasks remains underexplored, particularly for explicit termination modeling. Our work builds on CFG by integrating dynamic guidance with temporal step outputs, addressing the limitations of prior methods in modeling task progression and ensuring reliable termination in sequential robotic tasks.

\begin{figure*}[t]
\centerline{\includegraphics[width=\textwidth]{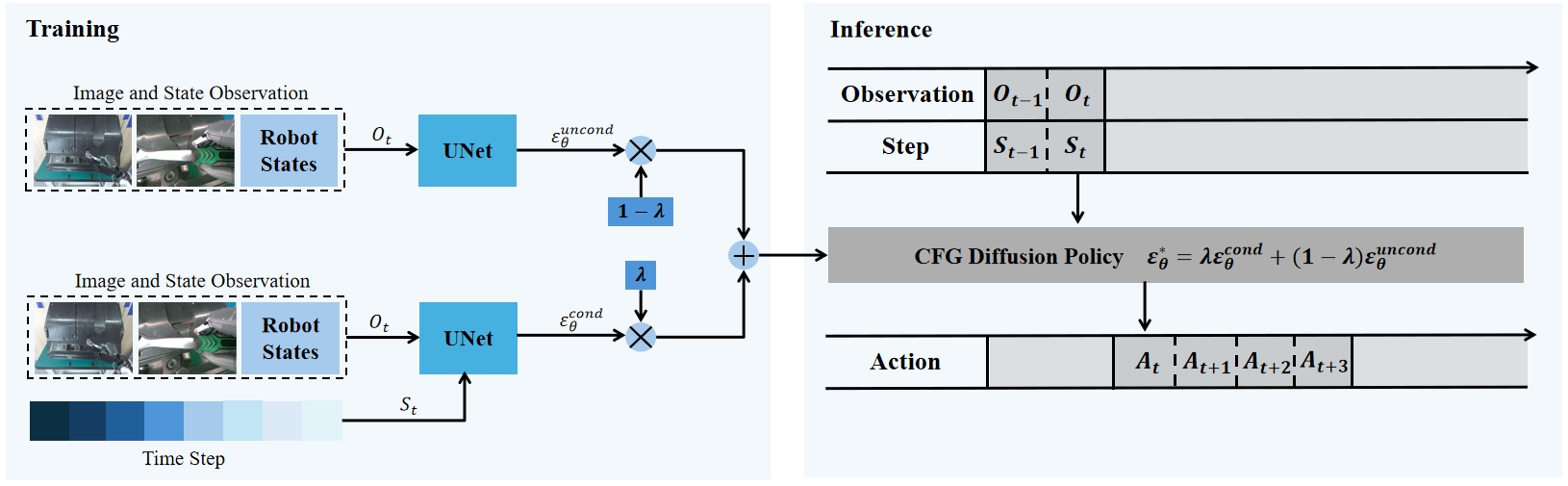}}
\caption{Overview of the proposed model architecture. \textit{Left:} Illustration of the observation processing pipeline, showing how different input are combined and processed. \textit{Right:} At time $t$, the policy processes the latest $T_o$ steps of observation $O_t$ and generates $T_a$ steps of actions $A_t$.}
\label{fig:model_overview}
\end{figure*}

\section{Methodology}
Our proposed framework enhances the DP by integrating 
CFG and explicitly modeling timestep 
counts \(S_t\) as input to improve termination 
condition judgment in temporal sequential robotic tasks. This section outlines the 
baseline DP framework, introduces our CFG-based policy with temporal modeling, and 
details the training and inference processes.

\subsection{Diffusion Policy Overview}
DP is a generative approach to visuomotor policy learning, 
leveraging DDPMs to map sensory inputs 
to action sequences~\cite{b1}. Given an observation \(\mathbf{O}_t\) at time \(t\),
 comprising visual inputs (e.g., RGB images from two cameras) and proprioceptive 
 states (e.g., arm motor states), DP predicts an action sequence 
 \(\mathbf{A}_t = [\mathbf{a}_t, \mathbf{a}_{t+1}, \ldots, \mathbf{a}_{t+T_a-1}]\) for \(T_a\) future steps. The policy models the conditional distribution \(p(\mathbf{A}_t | \mathbf{O}_t)\) using a diffusion process that iteratively denoises a noisy action sample.

The forward diffusion process starts with a clean action sequence \(\mathbf{A}_t^0 \sim p(\mathbf{A}_t | \mathbf{O}_t)\) and progressively adds Gaussian noise over \(K\) steps, defined as:
\begin{equation}
\mathbf{A}_t^k = \sqrt{\alpha_k} \mathbf{A}_t^0 + \sqrt{1 - \alpha_k} \boldsymbol{\epsilon}, \quad \boldsymbol{\epsilon} \sim \mathcal{N}(\mathbf{0}, \mathbf{I}),
\label{eq:forward_diffusion}
\end{equation}
where \(\alpha_k = \prod_{i=1}^k (1 - \beta_i)\), and \(\beta_i\) is a noise schedule parameter controlling the noise level at step \(i\). At \(k = K\), the distribution approximates \(\mathcal{N}(\mathbf{0}, \mathbf{I})\). The reverse process, which generates actions, is modeled as:
\begin{equation}
\mathbf{A}_t^{k-1} = \alpha \left( \mathbf{A}_t^k - \gamma \varepsilon_\theta(\mathbf{O}_t, \mathbf{A}_t^k, k) + \mathcal{N}(0, \sigma^2 \mathbf{I}) \right),
\label{eq:reverse_diffusion}
\end{equation}
where \(\varepsilon_\theta(\mathbf{O}_t, \mathbf{A}_t^k, k)\) is a neural network predicting the noise added at step \(k\), parameterized by \(\theta\), and \(\alpha, \gamma, \sigma\) are noise schedule functions~\cite{b1}. The training objective minimizes the mean squared error between predicted and actual noise:
\begin{equation}
\mathcal{L} = \mathbb{E}_{\epsilon, k} \left[ \|\epsilon - \varepsilon_\theta(\mathbf{O}_t, \mathbf{A}_t^0 + \epsilon, k)\|_2^2 \right],
\label{eq:dp_loss}
\end{equation}
where \(\epsilon \sim \mathcal{N}(\mathbf{0}, \mathbf{I})\) and \(k\) is sampled uniformly from \(1\) to \(K\). While effective for short-term action prediction, DP’s reliance on instantaneous \(\mathbf{O}_t\) limits its ability to track task progression, leading to poor termination judgment in cyclic tasks~\cite{b3}.

\subsection{Proposed CFG-Based Policy with Temporal Modeling}
To address the limitations of traditional DP, we introduce a CFG-based policy 
that incorporates current timestep as an input, enhancing deterministic control and 
reliable termination judgment. Our observation \(\mathbf{O}_t = [\text{visual}, 
\text{joint\_state}, \text{timestep}]\) includes a timestep count. The model 
predicts the action sequence \(\mathbf{A}_t\), using \text{timestep} to inform 
task progression for precise cycle completion.

\subsubsection{Classifier-Free Guidance}
Inspired by BESO~\cite{b5}, we employ CFG to guide action and timestep prediction toward termination actions at cycle endpoints. We train two models:
\begin{itemize}
    \item \textbf{Conditional Model}: Predicts the distribution \(p(\mathbf{A}_t, | \mathbf{O}_t, \mathbf{S}_t\)), conditioned on \(\mathbf{O}_t\) and \(\mathbf{S}_t\).
    \item \textbf{Unconditional Model}: Predicts \(p(\mathbf{A}_t | \mathbf{O}_t\)), ignoring {step}, to provide a baseline distribution.
\end{itemize}
During inference, CFG combines these predictions using a dynamic guidance factor \(\lambda\):

\begin{equation}
\begin{aligned}
\varepsilon_{\theta}^{*} = \lambda \varepsilon_{\theta}^{\text{cond}}(&\mathbf{O}_t, \mathbf{S}_t, \mathbf{A}_t^k, k) \\
&+ (1 - \lambda) \varepsilon_{\theta}^{\text{uncond}}(\mathbf{O}_t, \mathbf{A}_t^k, k),
\end{aligned}
\label{eq:cfg}
\end{equation}

where \(\varepsilon_{\text{cond}}\) and \(\varepsilon_{\text{uncond}}\) are noise predictions for the conditional and unconditional models, respectively. To emphasize termination as timestep approaches predefined threshold, we define:
\begin{equation}
    \lambda = \lambda_{\text{max}} \cdot \frac{1}{1 + e^{-(S_t - S_{t_0})}}
    \label{eq:lambda}
\end{equation}
where \(\lambda_{\max}\) sets the maximum guidance strength, \(S_t\) is the current timestep count, \(S_{t_0}\) represents the expected timestep at which the task nears completion.  The parameter \(S_{t_0}\) is derived from the average task length. This sigmoid function ensures that \(\lambda\) exhibits a gradual increase initially, accelerating as the cycle nears its conclusion, thereby enhancing the influence of the conditional model on termination actions~\cite{b5}.

\subsection{Model Architecture}
The proposed framework enhances the DP by integrating a CFG-based diffusion policy model to achieve robust action execution, as illustrated in Fig.~\ref{fig:model_overview}. The observation processing pipeline, shown in Fig.~\ref{fig:model_overview} (left), integrates multiple input modalities into a unified representation. At each timestep \( t \), the policy processes a sequence of observations \( \mathbf{O}_t = [\mathbf{O}_{t-T_o+1}, \ldots, \mathbf{O}_t] \), where \( T_o = 2 \) represents the observation horizon, capturing the latest two frames to model temporal dependencies. Each observation \( \mathbf{O}_i \) comprises three components: RGB images, proprioceptive joint states, and a timestep count, which tracks task progression as a scalar value.

In the observation processing pipeline, visual inputs are processed through a convolutional neural network (CNN) backbone, specifically a ResNet18~\cite{b1}, to extract spatial features. Concurrently, joint states and the timestep count are embedded into a shared feature space via separate linear embedding layers. The time
timestep count is mapped to a higher-dimensional embedding to ensure compatibility with other embeddings. These embeddings are concatenated to form the observation embedding for each frame, aggregated across the \( T_o \) frames to construct the input sequence \( \mathbf{O}_t \).

The CFG-based diffusion policy model, depicted in Fig.~\ref{fig:model_overview} (right), integrates action-sequence prediction with CFG to ensure temporal consistency and responsiveness. At timestep \( t \), the policy takes the observation sequence \( \mathbf{O}_t \) as input and predicts an action sequence \( \mathbf{A}_t = [\mathbf{a}_t, \mathbf{a}_{t+1}, \ldots, \mathbf{a}_{t+T_a-1}] \), consisting of 7-dimensional actions including 6 joints and dexterous hand. We define \( T_o \) as the observation horizon and \( T_a \) as the action execution horizon, balancing temporal action consistency with responsiveness to new observations. The model employs two denoising networks: a conditional model \( p(\mathbf{A}_t \mid \mathbf{O}_t, S_t) \), which uses the timestep count to guide task progression, and an unconditional model \( p(\mathbf{A}_t \mid \mathbf{O}_t) \), providing a baseline action distribution. During inference, CFG combines noise predictions from both models using a dynamic guidance factor \(\lambda\), as defined in \eqref{eq:lambda}, emphasizing termination actions as the task nears completion. The noise combination follows \eqref{eq:cfg}, ensuring guided action generation. The Denoising Diffusion Implicit Model (DDIM) sampler~\cite{b5} reduces the denoising process to 10 steps, achieving a latency of approximately 0.1 seconds on an Nvidia 3090 GPU.

This architecture leverages multi-frame observations and timestep count to enhance action coherence, while CFG ensures precise cycle completion, addressing limitations of traditional DP in cyclic tasks. The integration of visual, proprioceptive, and temporal inputs ensures deterministic control, as validated in real-world experiments.

\begin{figure*}[t]
  \centering
  \subfloat[DP Model\label{fig:subfigA}]{\includegraphics[width=\textwidth]{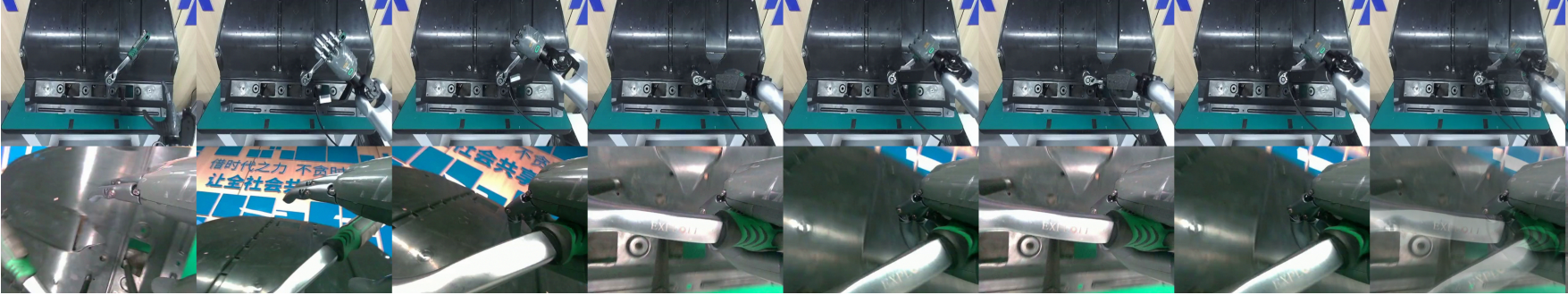}}
  \hfill
  \subfloat[CFG-DP Model (Ours)\label{fig:subfigB}]{\includegraphics[width=\textwidth]{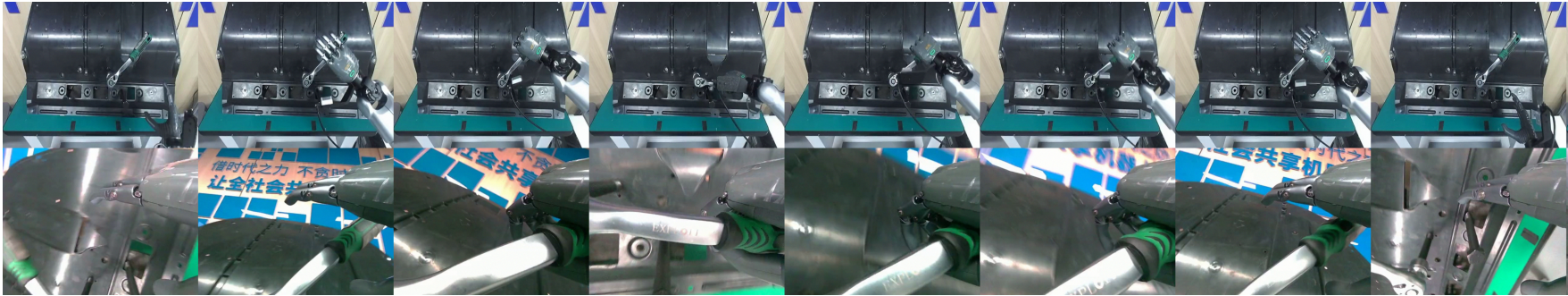}}
  \caption{Comparison of DP and CFG-DP models in the screwing task. (a) The DP model exhibits repetitive cyclic actions, failing to terminate properly. (b) The CFG-DP model successfully completes the task with a precise termination action.}
  \label{fig:main}
\end{figure*}

\section{Experiment}

\begin{figure}[tb]
\centerline{\includegraphics[width=1\columnwidth]{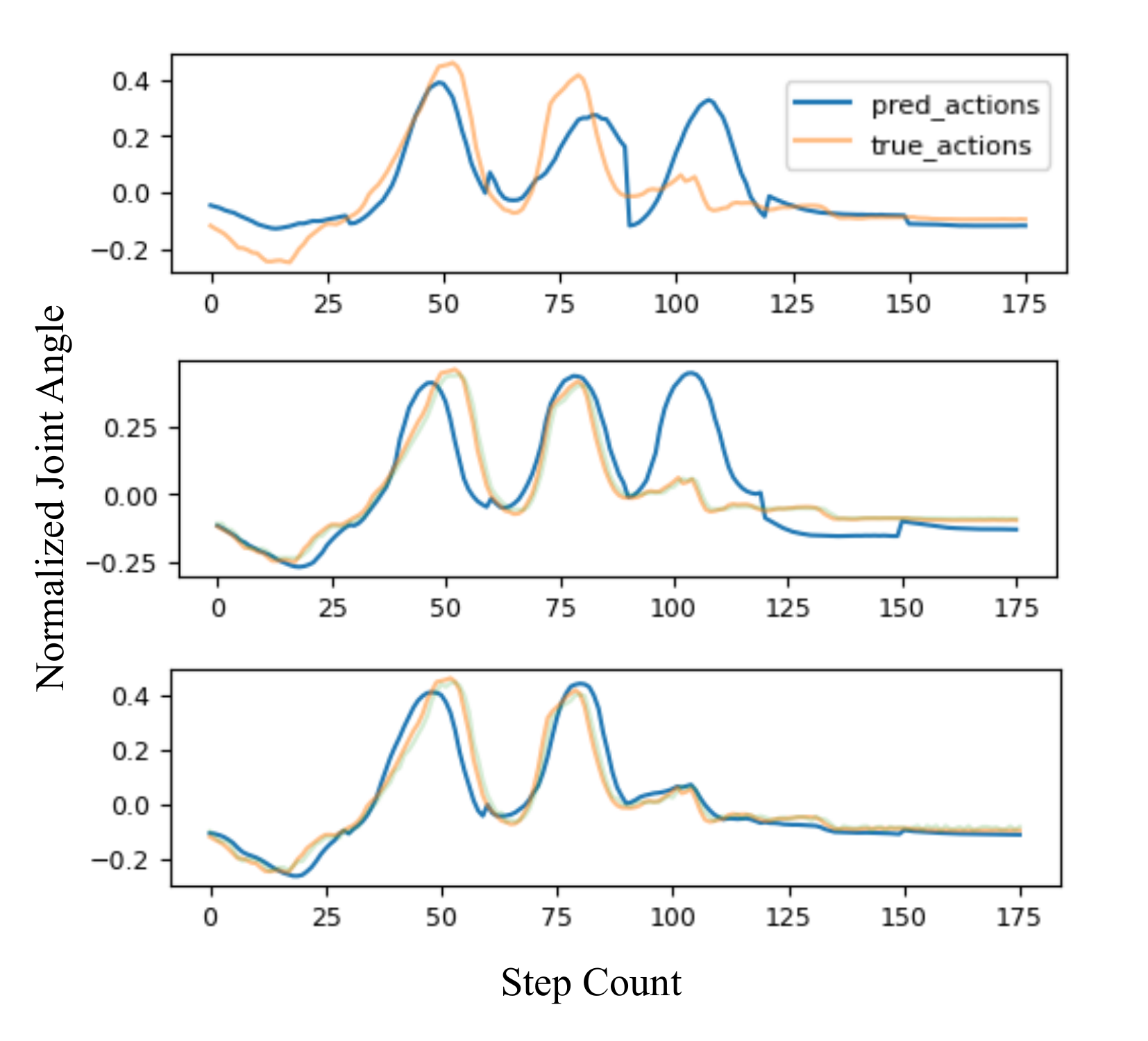}}
\caption{Action trajectories for the screwing task on the validation set, using wrist joint states. Top: Diffusion Policy (DP). Middle: Action Chunking with Transformers (ACT). Bottom: Classifier-Free Guidance Diffusion Policy (CFG-DP)}
\label{fig:state_response}
\end{figure}

The objective of our experiments is to verify and assess the feasibility and effectiveness of the proposed Classifier-Free Guidance Diffusion Policy (CFG-DP) with temporal modeling in a real-world humanoid robot screwing task. We evaluated CFG-DP against state-of-the-art baselines on a real-world screwing task and conducted ablation studies to analyze its design choices.

\subsection{Baselines}
We compare our CFG-DP against two state-of-the-art methods for robotic manipulation:

\begin{itemize}
  \item \textit{Diffusion Policy (DP)}: Uses a denoising diffusion model with a ResNet-18 backbone to generate up to 8-step action sequences. While effective for high-dimensional action spaces.
  \item \textit{Action Chunking with Transformers (ACT)}: Employs a Transformer to predict fixed-length action chunks, capturing short-term temporal dependencies.
\end{itemize}

All methods use position control with the BrainCo intelligent bionic dexterous hand, trained for 60,000 epochs with a batch size of 64, using the Adam optimizer. 

\subsection{Experimental Setup}
We evaluated the methods on a real-world screwing task, using a humanoid robot with 7-DoF arms and a BrainCo intelligent bionic dexterous hand. The task requires the robot to:
\begin{enumerate}
  \item Grasp a ratchet wrench with the dexterous hand.
  \item Perform three screwing cycles.
  \item Retract the arm to a designated end-zone.
\end{enumerate}

The environment includes two cameras: an Orbbec 335L (head, RGB + depth, 320$\times$240, 10 Hz) and a Realsense D405 (wrist, RGB + depth, 320$\times$240, 10 Hz). The observation space comprises RGB images, 7D joint angles, dexterous hand state, and timestep count. The action space consists of 7D actions, executed at 10 Hz without interpolation.

We collected 200 demonstration trajectories via VR-based teleoperation, each annotated with timestep counts. Trajectories included varied initial workpiece positions, two screwing cycles, and different termination positions to ensure robustness. The dataset was split into 80\% training and 20\% validation sets.

\subsection{Evaluation Metrics}
We assessed performance using three metrics:
\begin{itemize}
  \item \textit{Success Rate}: Percentage of rollouts completing two screwing cycles and retracting to the end-zone.
  \item \textit{Repetitive Actions}: Average number of redundant screwing rotations beyond the required two cycles per rollout.
  \item \textit{Completion Time}: Average time (in seconds) from rollout start to arm retraction, measured at 10 Hz.
\end{itemize}

\subsection{Performance on Validation Set}

We first evaluated the models on the validation set to assess their generalization to unseen initial conditions through qualitative analysis of action trajectories and end states, focusing on the wrist joint states, as shown in Figure~\ref{fig:state_response}. The figure illustrates the screwing task execution for DP, ACT, and CFG-DP across key events and final outcomes in simulation. For DP, the action trajectory reveals a tendency to get stuck in repetitive screwing cycles, failing to retract the arm properly, with the end state showing poor workpiece alignment. ACT exhibits even more pronounced issues, with jittery movements leading to excessive screwing cycles and no retraction to the end-zone; its end state indicates a complete misalignment of the workpiece. In contrast, CFG-DP demonstrates smooth and precise screwing cycles, completing the required two rotations and retracting the arm to the end-zone efficiently. The end state for CFG-DP shows the workpiece accurately aligned with the target, confirming successful task completion in simulation.

These qualitative results indicate that CFG-DP's temporal modeling and guided termination significantly improve generalization, enabling the policy to escape local optima and achieve reliable task execution. The promising performance of CFG-DP on the validation set justified its deployment for real-world experiments on the physical robot.

\subsection{Experimental Results on Humanoid Robot}

\begin{table}[tb]
  \centering
  \caption{Performance on the real-world humanoid robot screwing task.}
  \label{tab:results}
  \resizebox{\columnwidth}{!}{
    \begin{tabular}{lccc}
      \toprule
      \textbf{Method} & \textbf{Success Rate (\%)} & \textbf{Repetitive Actions} & \textbf{Completion Time (s)} \\
      \midrule
      DP & 55.6 & 2.6 & 32.0 \\
      ACT & 50.3 & 3.2 & 41.0 \\
      CFG-DP (Ours) & \textbf{83.2} & \textbf{0.3} & \textbf{24.2} \\
      \bottomrule
    \end{tabular}
  }
\end{table}

Following the validation, we conducted experiments on the physical humanoid robot to quantify task performance under real-world conditions. Table~\ref{tab:results} presents the results, showing that CFG-DP significantly outperformed DP and ACT across all metrics. CFG-DP achieved a success rate of 83.2\%, nearly eliminating repetitive actions with an average of 0.3 per rollout, and completed tasks in 24.2 seconds. In contrast, DP and ACT exhibited lower success rates respectively, and were prone to repetitive cycles.

\begin{figure}[tb]
\centerline{\includegraphics[width=1.1\columnwidth]{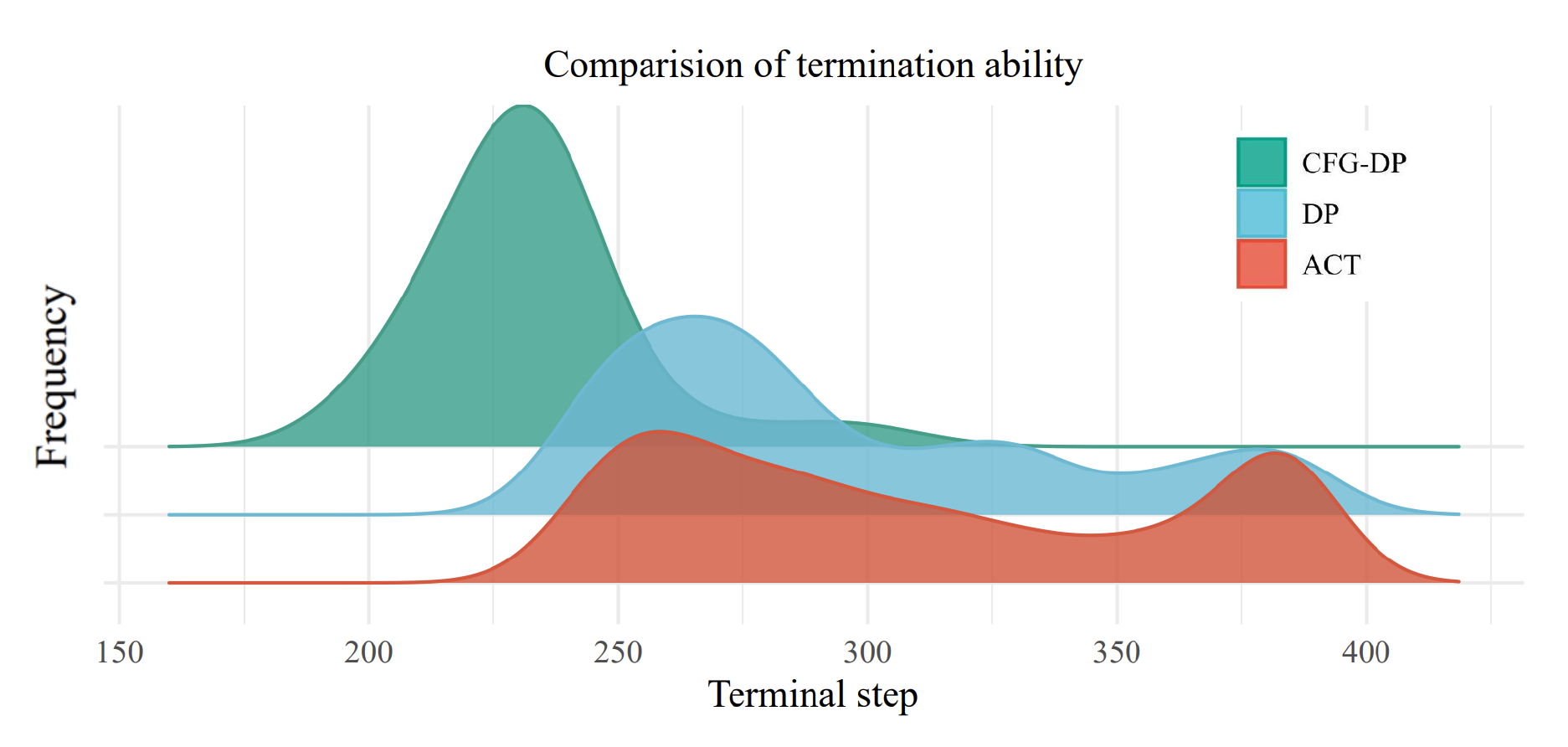}}
\caption{Distribution of termination steps for the screwing task.}
\label{fig:termination_dist}
\end{figure}

To further evaluate the termination behavior of CFG-DP, we analyzed the distribution of termination steps for CFG-DP alongside the baseline methods, DP and ACT. This comparison provides insights into how each method manages cycle completion and task termination, particularly in avoiding repetitive actions.

Figure~\ref{fig:termination_dist} illustrates the termination step distributions for CFG-DP, DP, and ACT. The baseline methods, DP and ACT, show broadly similar distributions, with means shifted toward higher timestep counts and extended right tails reaching well beyond the expected termination point. This wider spread reflects their shared difficulty in accurately determining when to end the task without explicit temporal cues. Both methods often engage in excessive screwing cycles, leading to delayed task completion, as observed in their tendency to produce more repetitive actions and longer overall execution times. This behavior aligns with their qualitative performance in Figure~\ref{fig:state_response}, where both model results in redundant rotations.

In contrast, CFG-DP's distribution is notably more concentrated, with a mean closer to the expected terminal step and a sharply compressed right tail, resembling a truncated normal distribution. This tighter distribution highlights CFG-DP's ability to reliably converge on the appropriate termination step after completing the required screwing cycles. The integration of timestep counts and dynamic $\lambda$ enables CFG-DP to effectively minimize repetitive actions. The concentrated distribution of CFG-DP, compared to the broader distributions of DP and ACT, underscores its superior termination precision in cyclic tasks.

\subsection{Ablation Study}

\begin{figure}[tb]
\centerline{\includegraphics[width=1\columnwidth]{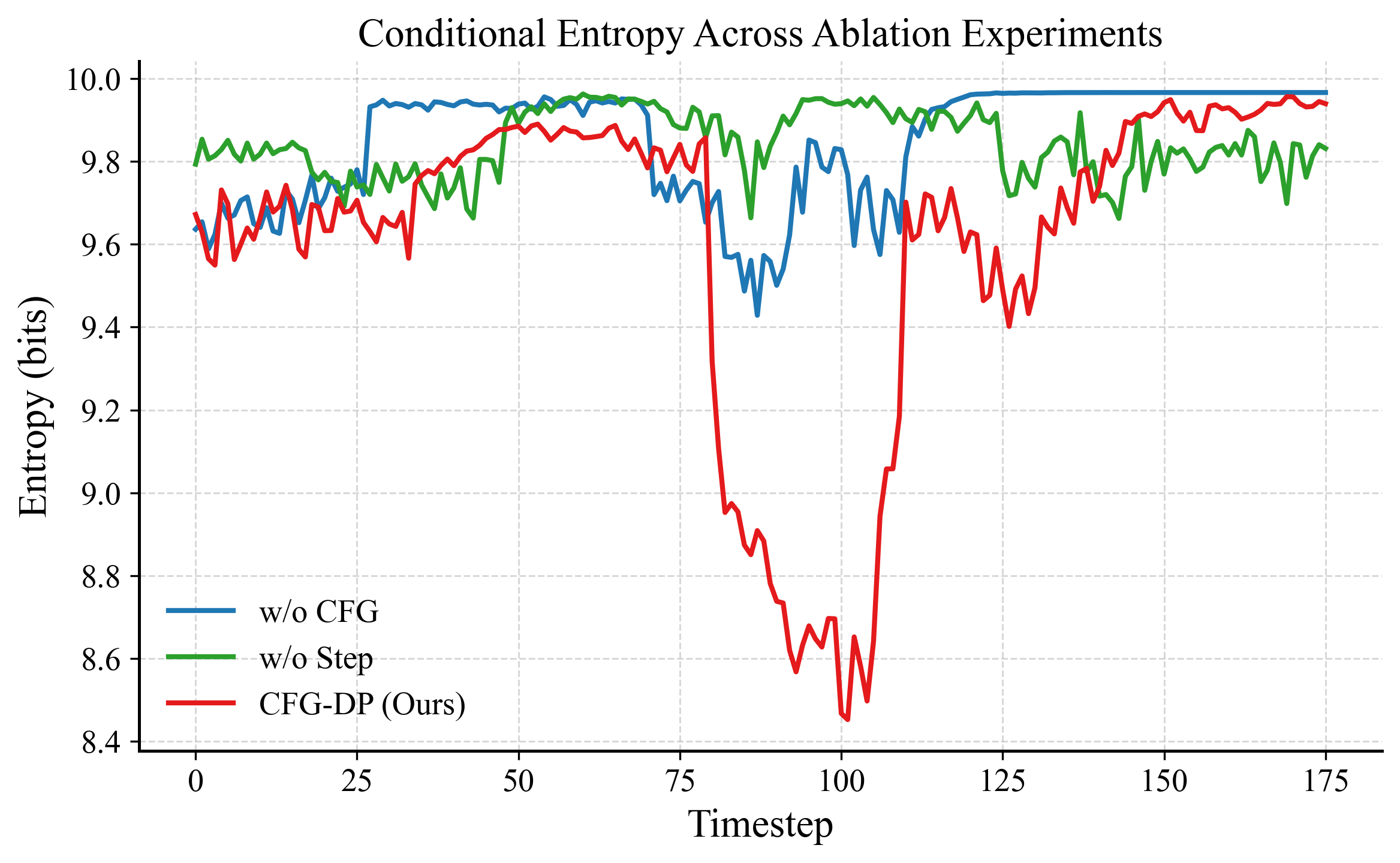}}
\caption{Conditional entropy across ablation experiments}
\label{fig:conditional_entropy}
\end{figure}

\paragraph{Qualitative Analysis}
To assess the contributions of CFG, timestep input \( S_t \), and dynamic guidance factor \( \lambda \) in our CFG-DP, we conducted ablation studies. Experiments were performed on the real-world screwing task described in Section IV.B. Our analysis focuses on the qualitative evolution of conditional entropy, providing insights into the policy's action distribution dynamics across three configurations: (1) without CFG (w/o CFG), (2) without timestep input (w/o Step) and (3) the complete CFG-DP model (Ours).

To investigate behavioral differences across configurations, we analyzed the conditional entropy \( H(\mathbf{A}_t | \mathbf{O}_t) \), which measures the uncertainty in the policy's action distribution at each timestep. This metric is derived from 100 Monte Carlo samples, with the 7D action space discretized into 100 bins to estimate the distribution.

Fig.~\ref{fig:conditional_entropy} illustrates the temporal evolution of conditional entropy, offering key insights into the policy's behavior across ablation experiments. While all three configurations exhibit an entropy reduction near termination timesteps, suggesting that the underlying data distribution—where demonstration trajectories converge toward a withdrawal action—may inherently drive some degree of determinism. CFG-DP significantly amplifies this effect, achieving a more pronounced and earlier convergence due to its adaptive guidance. 

In comparison, the w/o CFG configuration maintains a persistently high entropy, indicating a lack of effective exploration-exploitation balance without CFG, resulting in oscillatory and suboptimal action patterns. Similarly, the w/o Step configuration shows elevated entropy with significant fluctuations, as the absence of \( S_t \) needed for CFG to function optimally. Despite identical inputs for conditional and unconditional models in this case, CFG retains a residual role by leveraging the distinct distributions learned during training through conditional dropout, though its impact is limited compared to the complete model.

The CFG-DP configuration demonstrates a marked reduction in conditional entropy near the termination step. This sharp decline reflects the policy's ability to converge to a highly deterministic action distribution, aligning with the task's requirement for precise withdrawal following the screwing cycles. The dynamic adjustment of \( \lambda \), enhances the influence of the conditional model. This mechanism suppresses the exploratory tendencies of the unconditional model, focusing the action set and improving decision precision at the termination point. Additionally, the model will still maintain high conditional entropy at other timesteps to ensure multimodality. Overall, CFG-DP modal leverages the diffusion model's multi-modal capacity while adapting to the task's phased requirements.

\paragraph{Quantitative Analysis}

\begin{figure}[tb]
\centerline{\includegraphics[width=0.9\columnwidth]{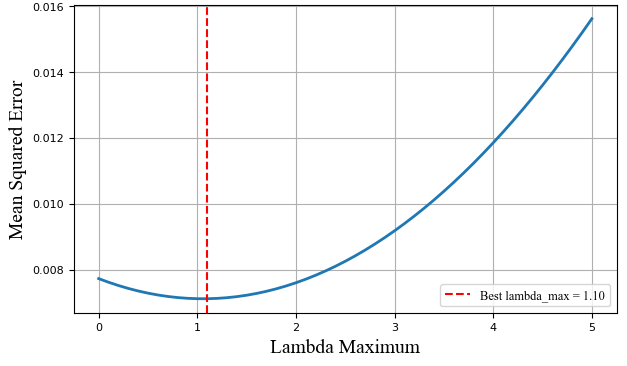}}
\caption{Ablation study on the effect of \(\lambda_{\text{max}}\) on prediction accuracy. Plot of Mean Squared Error (MSE) across \(\lambda_{\text{max}}\) values from 0 to 5, showing the lowest MSE at \(\lambda_{\text{max}} = 1.10\).}
\label{fig:lambda_max_mse}
\end{figure}

To assess the effect of the maximum regularization parameter \(\lambda_{\text{max}}\) on our model's performance, we conducted an ablation study by varying \(\lambda_{\text{max}}\) from 0 to 5 and evaluating its impact on prediction accuracy via Mean Squared Error (MSE). Each configuration was averaged over 10 independent trials.

These results highlight the importance of parameter tuning, with \(\lambda_{\text{max}} = 1.10\) achieving the best performance, as evidenced in Figure~\ref{fig:lambda_max_mse}. We speculate that this value optimally balances model complexity and extreme parameter values adversely affect action prediction outcomes, as confirmed by the MSE trend.

\section{Conclusion}
This work introduces a novel framework to enhance humanoid robot performance 
in temporal sequential tasks, effectively mitigating repetitive actions 
that challenge traditional visuomotor policies. By integrating 
CFG with conditional and unconditional models 
and leveraging timestep counts for temporal modeling, our approach 
greatly improve the decision-making ability of tranditional DP model to terminate actions.
Throughout experiments on both benchmark datasets and a humanoid robot platform, we we have demonstrated a significant improvement in termination success rate in screwing test.
Ablation studies further elucidate the framework’s strengths: optimal tuning of \(\lambda\) ensures highest prediction accuracy, conditional entropy analysis demonstrates deterministic action convergence at termination while preserving multimodal exploration. These findings highlight the pivotal role of temporal modeling and CFG in overcoming local optima traps inherent in baseline methods.

This framework ensures robust deterministic control and execution reliability, 
outperforming baseline policies through precise task termination. 
However, sensitivity to timestep count estimation persist as challenges. Future work will focus on developing more robust models, such as through reinforcement learning to optimize timestep count prediction and reduce reliance on manual parameter tuning.
Additionally, while the framework focuses on screwing tasks, future work will develop hierarchical or modular strategies to decompose complex tasks into sub-tasks, enabling multi-cycle operations like assembly or pick-and-place, and investigate multi-task learning to adapt policies across diverse operations without retraining.
This advancement achieve
deterministic control and reliable execution in temporal sequential tasks, with potential to transform autonomous 
robotic operations.

\section*{Acknowledgment}

We sincerely thank the reviewers for their valuable time and insightful feedback, which greatly contributed to improving the quality of this work. This research was supported by the Shenzhen Special Fund for Future Industrial Development (No. KJZD20230923114222045).

\end{document}